%% file: 0_main.tex
\documentclass[letterpaper, 10 pt, conference]{ieeeconf}  

\IEEEoverridecommandlockouts                              

\usepackage{graphics} 
\usepackage{mathptmx} 
\usepackage{amsmath} 
\usepackage{amssymb}  
\usepackage{graphicx}
\usepackage{subcaption}
\input{macros.tex}

\title{\LARGE \bf
Soft Hybrid Aerial Vehicle via Bistable Mechanism
}

\author{Xuan Li$^{*}$, Jessica McWilliams$^{*}$, Minchen Li, Cynthia Sung, Chenfanfu Jiang
\thanks{$^{*}$ Equally contributed}
\thanks{X. Li, M. Li, and C. Jiang are with the SIG Center for Computer Graphics, University of Pennsylvania (emails: \{xuanli1, minchenl, cffjiang\}@seas.upenn.edu). J. McWilliams and C. Sung are with the General Robotics, Automation, Sensing \& Perception (GRASP) Lab, University of Pennsylvania (emails: \{jmcw, crsung\}@seas.upenn.edu). Support for this project has been provided in part by NSF Grant No. 1138847, 1943199, 1813624, 2023780 and DGE-1845298. We would also like to thank Jason Friedman for CAD assistance, Yifan Yuan for 3D printing the TPU arms, and Terry Kientz and Jeremy Wang for their assistance with fabricating the HAV and rotating boom.}
}

\begin{document}

\maketitle

\begin{abstract}
Unmanned aerial vehicles have been demonstrated successfully in a variety of tasks, including surveying and sampling tasks over large areas. These vehicles can take many forms. Quadrotors' agility and ability to hover makes them well suited for navigating potentially tight spaces, while fixed wing aircraft are capable of efficient flight over long distances. Hybrid aerial vehicles (HAVs) attempt to achieve both of these benefits by exhibiting multiple modes; however, morphing HAVs typically require extra actuators which add mass, reducing both agility and efficiency. We propose a morphing HAV with folding wings that exhibits both a quadrotor and a fixed wing mode without requiring any extra actuation. This is achieved by leveraging the motion of a bistable mechanism at the center of the aircraft to drive folding of the wing using only the existing motors and the inertia of the system. We optimize both the bistable mechanism and the folding wing using a topology optimization approach. The resulting mechanisms were fabricated on a 3D printer and attached to an existing quadrotor frame. Our prototype successfully transitions between both modes and our experiments demonstrate that the behavior of the fabricated prototype is consistent with that of the simulation. 

\end{abstract}

\input{1_Introduction}
\input{3_formulation}

\input{4_evaluation}

\input{5_discussion}




\bibliographystyle{IEEEtran}
\bibliography{IEEEabrv,reference}

\end{document}

%% file: macros.tex
\usepackage{xcolor}

%% file: 1_introduction.tex
\section{INTRODUCTION}

Hybrid aerial vehicles (HAVs) aim to improve flight efficiency and vehicle versatility by embedding in a single vehicle the ability to achieve multiple flight modes~\cite{ke2018design,becker2016designing,barbarino2011review}: a  maneuverable copter mode capable of vertical takeoff and landing, hover, and other agile maneuvers; and a fuel-efficient fixed wing mode aimed at long-distance flight.
Morphing aerial vehicles achieve different flight modes by changing the morphology of the vehicle itself~\cite{falanga2018foldable,zhao2018dragondrone,zhao2018transformable}, but such morphing behavior often incurs additional costs of added weight, complexity and control~\cite{xu2019learning,floreano2017foldable,tan2020morphable,morton2017small}.

In this paper, we propose a new HAV design (Fig.~\ref{fig:full_hav}), wherein the vehicle switches between a quadrotor mode and a fixed wing mode via a compliant bistable mechanism that deploys wings without requiring any additional actuators for reconfiguration beyond those included for normal flight. 
The design is inspired by \cite{bucki2019design}, in which the arms of a quadrotor fold inward when the thrust is below a certain threshhold to allow the vehicle to fit through a tight space.
We replace the folding mechanism with a bistable mechanism, allowing the HAV to remain in either mode in the absence of thrust. Our system replaces rigid quadrotor arms with soft material which deforms between two stable configurations, but constrains the propellers to a fixed ring. The first stable mode causes wings to deploy, and the other mode folds the wings. To transition between modes, we show that a sudden change in thrust can cause the battery to lag behind the motion of the outer ring due to its inertia, and that this inertial lag can be used to generate a mode switch without requiring any extra actuators. For this concept to work, the force required to actuate the bistable mechanism must fall within a specified range, based on the mass and thrust capabilties of the system.

\begin{figure}[tb]
    \centering
    \includegraphics[width=0.4\textwidth]{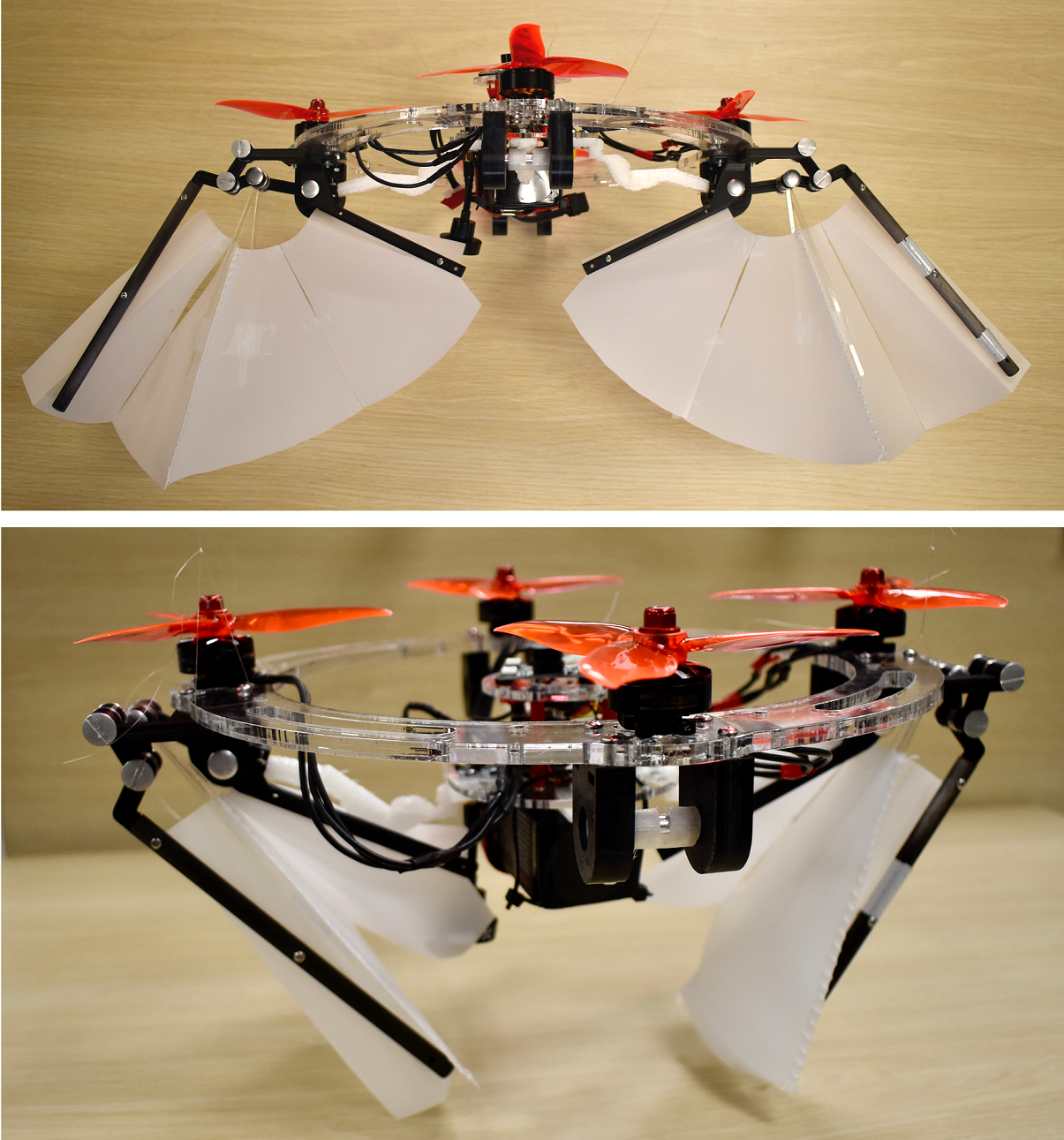}
    \caption{The fabricated HAV prototype with fixed wing (top) and quadrotor (bottom) modes.}
    \label{fig:quad_proto}
    \vspace{-0.5cm}
\end{figure}

Our vehicle is automatically optimized via  \emph{differentiable elasticity simulation}, where the mode switching mechanism is optimized by differentiating the material topology to achieve a compliant bistable mechanism and the wing deployment mechanisms is designed by differentiating the multi-bar connectivity and its reachability.
Using topology optimization to design bistable structures has been studied for years. Most previous work rely on nonlinear finite element analysis to achieve large deformations, where robustly solving the displacement control and computing the analytic sensitivity information remain challenging. \cite{Prasad2005} maximized the distance between two equilibrium states. But they used a generic algorithm to optimize the topology without sensitivity information, which is not efficient. \cite{James2016} achieved bistability by manipulating the force-displacement curve directly. They minimized the backward switching force and set a lower-bound for the forward switching force. Arc-length method was utilized to achieve displacement control. \cite{Chen2018} followed this formulation, but achieved displacement control by increasing a prescribed displacement gradually to a target one. \cite{Chen2019} proposed a new formulation that optimizes the range of the force-displacement curve so that two-direction switching forces can both get optimized. We follow this idea as part of our formulation. We further model the displacement control as an energy minimization problem under equality constraints. With augmented Lagrangian method and project Newton method \cite{teran2005robust}, our approach is conceptually simple and practically robust even with large deformations.

We further utilize a multi-body mechanism for the wing design to enable large rotations. Our idea stems from \cite{Swartz2019}, which connects layers of meshes with clusters of springs with decayed stiffness to simulate pin joints between layers. The major advantage of this model is in the effective enforcement of positional continuity. Inspired by this work, we model joints as equality constraints between the interpolated displacements on the same material coordinate of two different layers in our optimization-based equilibrium solver. Consequently, each joint is placed inside a computational cell with its position optimized.

The contributions of this paper include:
\begin{itemize}
\item a novel soft Hybrid Aerial Vehicle (HAV) that leverages a compliant bistable mechanism for achieving two flight modes: 1) a quadrotor mode enabling maneuverability and 2) a fixed wing mode enabling efficient flight without requiring additional actuators;
\item a topology optimization approach for both the bistable switching mechanism and the wing deployment mechanisms of the HAV; and
\item experimental validation of the HAV design in a fabricated prototype.
\end{itemize}

%% file: 3_formulation.tex
\section{Design and Validation of the HAV System with Differentiable Simulations}
The automated design pipeline for our HAV system is materialized using efficient simulations of both elastostatics and elastodynamics. First, with a \emph{differentiable} deformable body simulator that solves for static force equilibrium, the design task is formulated as a smooth optimization problem where design variables are optimized with sensitivity information back-propagated from the objective (Sec.\ \ref{sec:bistableTopoOpt} and\ \ref{sec:multiBodyDesign}). Second, the elastodynamic simulation quickly validates each design and filters out inferior design choices before fabrication (Sec.\ \ref{sec:validation}). 

The HAV system includes a central bistable structure and a foldable wing structure, wherein the state transition of the bistable structure folds wings inside or opens them up. These two parts are sequentially designed: a nonlinear elastic  topology is first optimized to obtain bistability, and then the rotations of the joints of its arms act as boundary conditions for the wing design, where we reuse the differentiable equilibrium solver to optimize a multi-body rigid mechanism and obtain optimal joint locations for the wings.

\subsection{Nonlinear Topology Optimization for Soft Materials}
Topology optimization tackles the inverse simulation problem of finding a material distribution that fulfills mechanical and geometrical requirements under static equilibrium. We perform topology optimization on nonlinear hyperelastic materials to design the central bistable mechanism. Here we review the adopted topology optimization machinery.

Density-based topology optimization\ \cite{andreassen2011efficient} usually represents a structure with a smooth density field  $\rho \in [0,1]$ in the material space $\Omega$, where $1$ represents fully solid and $0$ represents fully void. Then topology optimization can often be desribed as a constrained optimization problem with the objective function $L$ being the elastic potential, or compliance, and the constraints specifying static force equilibrium condition and the material volume target. 

We choose the neo-Hookean  hyperelasticity\ \cite{bonet1997nonlinear} to model nonlinear elastic deformations that are crucial for bistable transitions.
We then adopt a differentiable Material Point Method (MPM) \cite{jiang2016material,hu2019chainqueen} for discretization. MPM is a hybrid Lagrangian-Eulerian approach for computational solids, where the deformation is discretized on quadrature particles and physical equations are discretized on a grid.
The sensitivity information is computed by differentiating the objective and the volume constraint with respect to the design variables\  \cite{li2020leto}.
With gradients we solve the optimization problem using the Method of Moving Asymptotes (MMA) \cite{svanberg1987MMA}.
 
\subsection{Bistable Mechanism via Topology Optimization}\label{sec:bistableTopoOpt}
Bistable mechanism allows elastic structures to contain two stable equilibrium states, both of which can stably maintain their shapes without requiring any external loads. It is especially suitable for designing HAVs with two modes.

\begin{figure}
    \centering
    \begin{subfigure}{0.26\textwidth}
    \includegraphics[width=\textwidth]{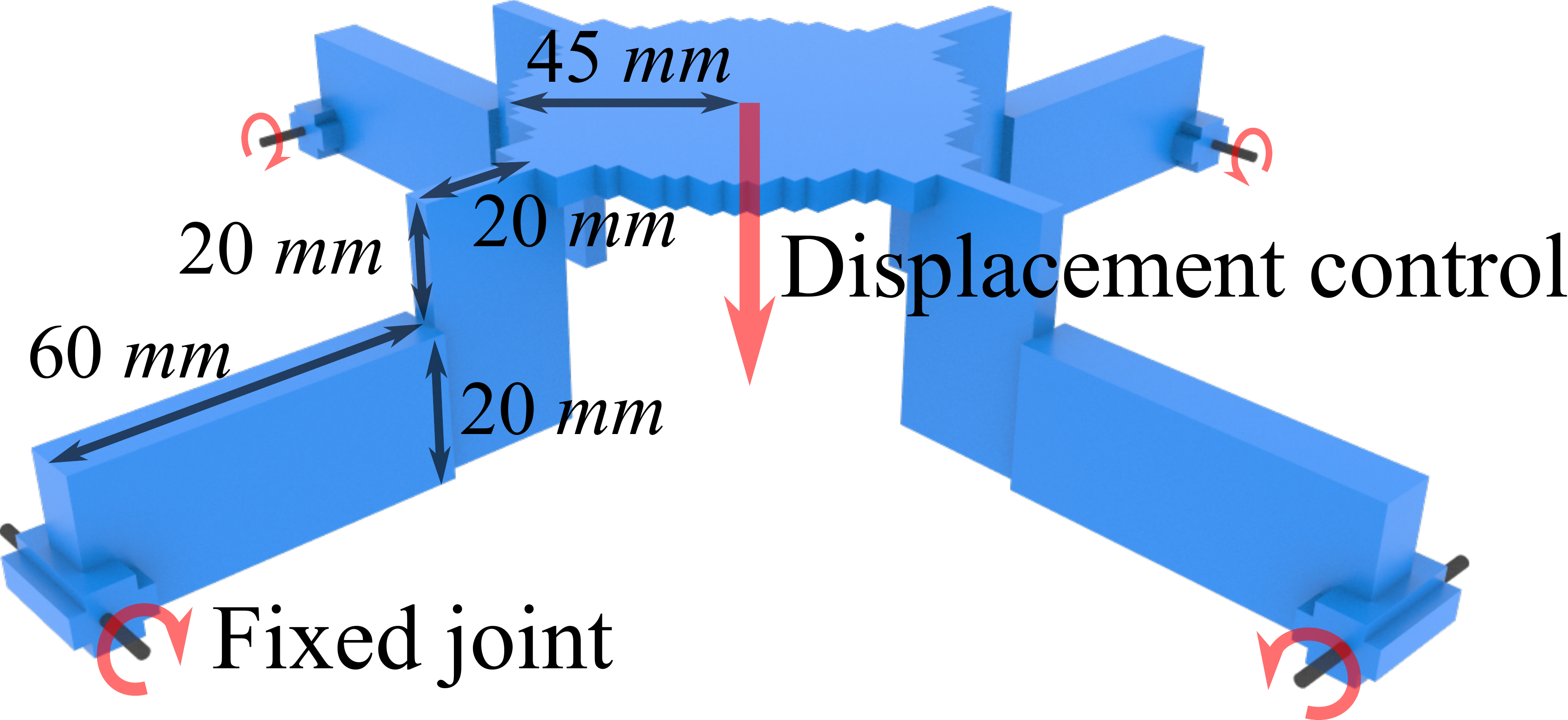}
    \caption{}
    \label{fig:bistable_initial}
    \end{subfigure}
    \begin{subfigure}{0.2\textwidth}
    \includegraphics[width=\textwidth]{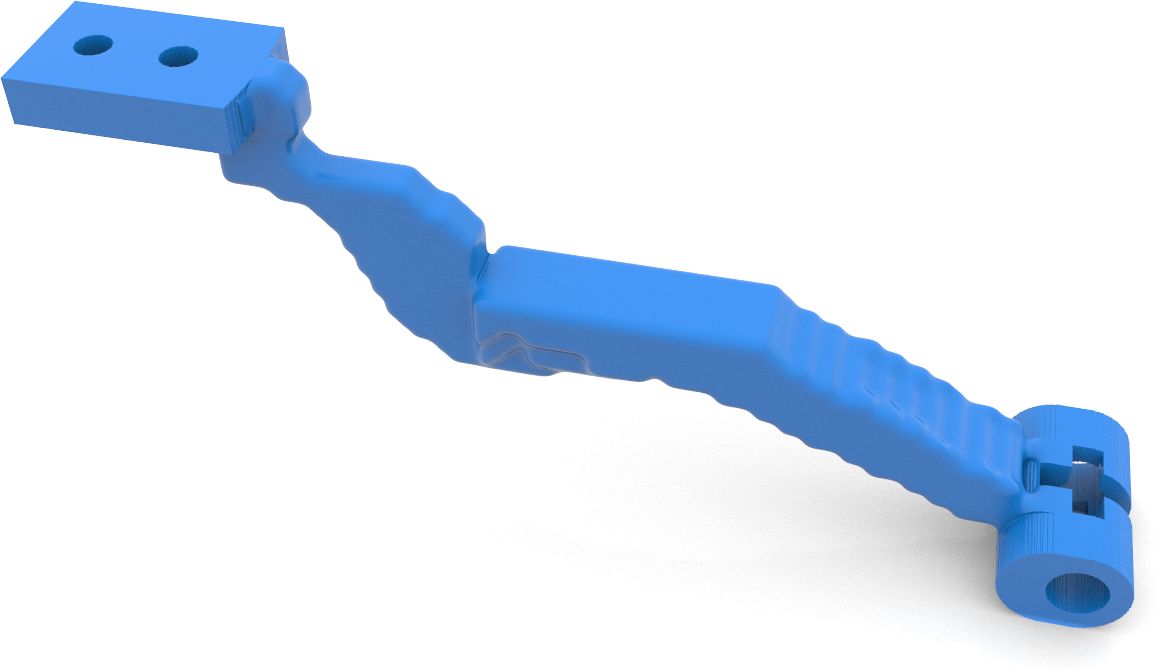}
    \caption{}
    \label{fig:optimized_arm}
    \end{subfigure}
    \caption{(a) The design configuration, where the design domain of each arm is of an L shape, and the control direction is visualized with the red arrows. Each arm can rotate around an axle on its end; (b) The final optimized arm.}
    \label{fig:bistable_optimization}
    \vspace{-0.5cm}
\end{figure}

Following existing literature \cite{James2016,Chen2018,Chen2019}, we tackle bistable mechanism by controlling the force-displacement curve. The structure is bistable when the curve spans above and below the $f=0$ (zero force) line. The force-displacement curve is acquired through displacement control in the quasi-static setting: certain ports are constrained on a prescribed path by a given displacement sequence, then the reaction force on a port (the force needed to maintain the port on the track) is computed sequentially by static equilibrium on other nodes. See Fig.  \ref{fig:bistable_initial} for our choice of ports. This quasi-static problem at each control point $i$ can be modelled as an energy minimization problem subject to an equality constraint:
\begin{equation}
        \min_{u}\ e(\rho, u) \ \ \ \text{s.t.}\ \   u_i = u^*_i
    \label{eq:pure-dirichlet}
\end{equation}
where the displacement of node $i$ is prescribed to some non-zero value $u_i^*$. Note that fixed nodes with zero displacements are eliminated from the degrees of freedom directly. The equality constraint is handled using the augmented Lagrangian method. With project Newton and non-invertible line search \cite{li2020leto,nocedal2006numerical}, the solver remains robust under arbitrarily large displacement-control constraints. After the displacement field is acquired, the reaction force at the port $i$ is then computed as $R_i = \frac{\partial e}{\partial u_i}$, which is also the Lagrangian multiplier for the equality constraint at the optimal point.

Similarly to \cite{Chen2019}, we maximize the difference of two switching forces (the forces required for snap-throughs between the two states), while minimizing the mean compliance under a force along the control path to guarantee sufficient structural stiffness. The complete formulation of the bistable mechanism topology optimization is then 
\begin{equation}
\begin{split}
    &\min_{\xi}\ n^T\left(\frac{\partial e(\rho, u^2)}{\partial u_i} - \frac{\partial e(\rho, u^1)}{\partial u_i} \right) + \alpha f^T u^3\\
    \text{s.t.} \quad
&\begin{cases}
    u^1 = \operatornamewithlimits{argmin}_u e(\rho, u)\ \  \text{s.t.} \ u_i = \bar{u}_i^{1}\\
    u^{2} = \operatornamewithlimits{argmin}_u e(\rho, u)\ \  \text{s.t.} \ u_i = \bar{u}_i^{2}\\
    u^3 = \operatornamewithlimits{argmin}_u e(\rho, u) - u^T f \\
     n^T\frac{\partial e(\rho, u^1)}{\partial u_i} \le f^*_1, \ \ n^T\frac{\partial e(\rho, u^2)}{\partial u_i} \ge - f^*_2\\
    V(\rho) \le \bar{V},
\end{cases}
\end{split}
\label{eq:bistability}
\end{equation}
where $n$ is the control path direction, $f$ is the regularity force along $n$, and $\alpha$ controls the weighting between two objectives. $\bar{u}_i^{1}$ and $\bar{u}_i^{2}$ correspond to our expected peak and valley points in the force-displacement curve. $f^*_1$ and $f^*_2$ are used to control the magnitude upper bound of snap-through forces to match practical needs. 

The derivative of the reaction force $R_i = \frac{\partial e}{\partial u_i}$ w.r.t design variable $\xi$ contains term $\frac{d\widehat{u_i}}{d\rho}$, which can be acquired by differentiating the force equilibrium equation $ \frac{\partial e}{\partial \widehat{u_i}} = 0$ on $\widehat{u_i}$ w.r.t $\rho$. This leads to
\begin{equation}
    \frac{d R_i}{d \xi} = \left[\frac{d\rho}{d\xi}\right]^T \left(\frac{\partial e^2}{ \partial \rho \partial u_i} +  \frac{\partial^2 e}{\partial \rho \partial \widehat{u_i}} \left[ \frac{\partial^2 e}{\partial \widehat{u_i}^2} \right]^{-1} \frac{\partial^2 e}{\partial \widehat{u_i} \partial u_i} \right).
\end{equation}

\begin{figure}
    \centering
    \includegraphics[width=0.4\textwidth]{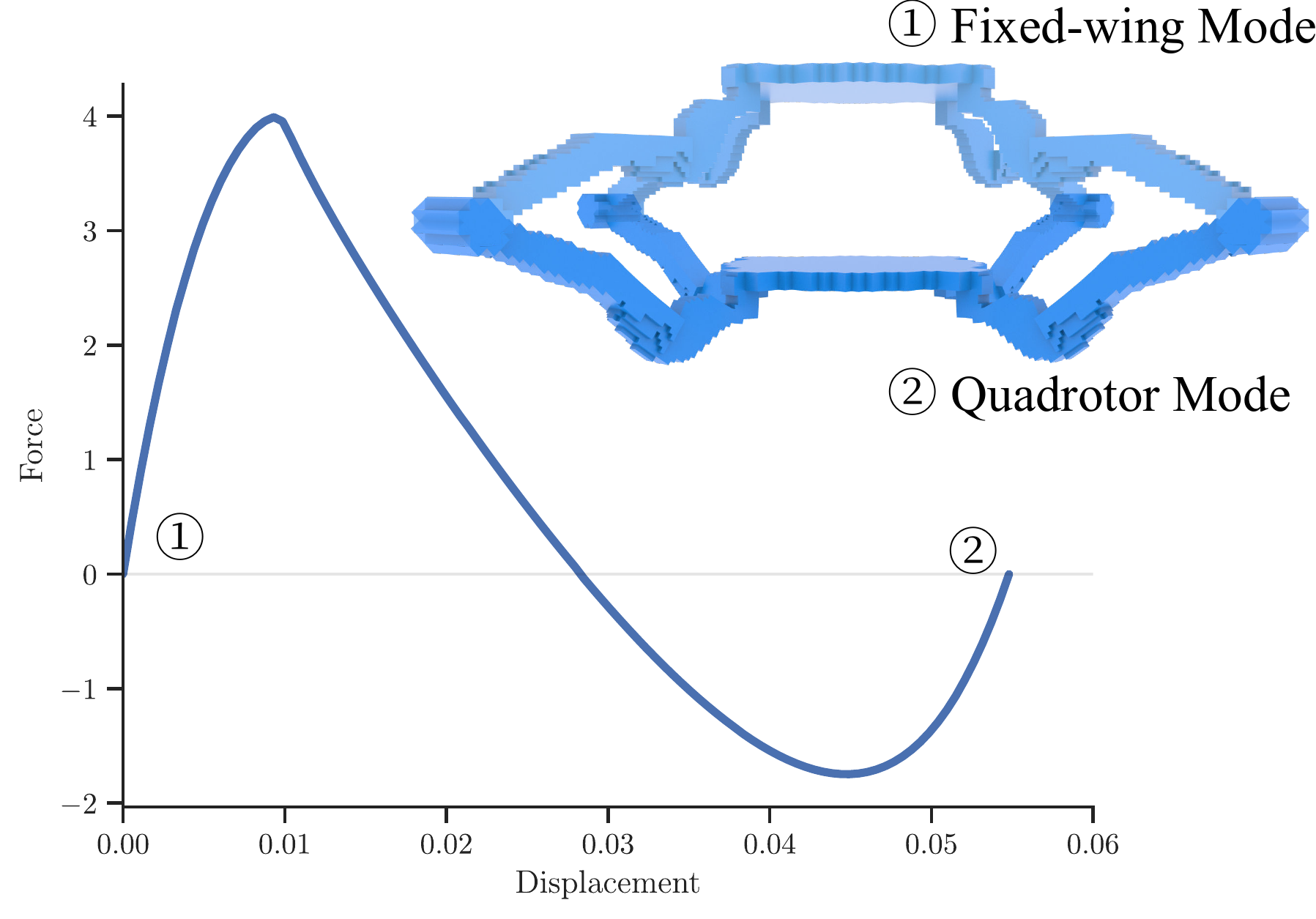}
    \caption{Force-displacement plot of the optimized topology with both equilibrium configurations visualized.}
    \label{fig:two-stables}
    \vspace{-0.5cm}
\end{figure}

The settings of the initial design domain, the control direction and the Dirichlet boundary condition are illustrated in Fig. \ref{fig:bistable_initial}. We model the initial topology following the configuration of a quadrotor, and initialize the density in the design domain to be the target volume fraction everywhere. Each L shape is constructed by the constructive solid geometry difference between a box of 80 mm $\times$ 40 mm and a small box of 60 mm $\times$ 20 mm. This layout effectively avoids potential collisions between the arm and other parts on the quadrotor. The arm width is 5 mm for the first 20 mm close to the center and 10 mm for the other part. Each such domain eventually becomes an arm connecting the big ring with radius 125~mm to a central housing with radius 45 mm. The central plate represents the inner housing platform whose density is fixed to be 1 throughout the optimization. In addition to the displacement control constraints, the displacements of the axles (represented with small cylinders) are also fixed to be zero so that the arms can rotate around them (visualized with gray bars in Fig.  \ref{fig:bistable_initial}). 
For efficiency, we only optimize over a quarter of the whole domain. Force equilibrium is solved under two symmetric boundary conditions so that the displacement field coincides with the one computed with the full domain. Likewise, to ensure identical arms, we set an extra symmetry constraint on the density field of each arm. 
The base Young's modulus and Poisson ratio of the arm are set as $5.5 \times 10^6$ Pa and $0.48$ (TPU's material parameters). The Young's modulus of the central plate is set to be $1000$ times larger (thus essentially treated as a rigid body). The expected peak and valley points of the force-displacement curve are chosen as $\bar{u}_i^1 = 10$ mm and $\bar{u}_i^2 = 50$ mm. Other parameters include $f_1^* = 4$ N, $f_2^* = 1.2$ N, $f = -f_1^*n$, $\bar{V} = 0.2$, and $\alpha=50$. The force parameters are chosen according to the available mechanical parts of the quadrotor. The optimized single arm is shown in Fig. \ref{fig:optimized_arm}. The full structure with four arms (Fig.  \ref{fig:two-stables}) demonstrates two equilibrium states.

\subsection{Multi-Body Mechanism for the Wing Design}
\label{sec:multiBodyDesign}
The folding of the wing is driven by the state transition of the bistable structure. One potential design strategy is to use the compliant mechanism \cite{Pedersen2001}. However, the rotation of the wing blade is much larger than the arm of the bistable structure, and in this situation compliant mechanism tends to generate low-density elements to act as joints, which makes the fabrication challenging. Inspired by \cite{Swartz2019}, we use multiple pieces of the continuum material to represent different components and use pinned joints (points in 2D and segments in 3D) to connect them. The material spaces of all pieces are aligned so that the same material coordinate refers to the same position in world coordinate at the undeformed state. We model a joint as one (for 2D joints) or a set of (for 3D joints) equality constraint(s) in the form of $u^1(X_i) = u^2(X_i)$ when solving the force equilibrium, where $u^1$ and $u^2$ are displacement fields of two different pieces that we connect and $X_i$ is the common material coordinate. We only need a foldable wing skeleton, so we set the topology as the simplest form: we connect joint positions on the same component directly by straight bars. The only optimization variables are then the positions of these movable joints.

For any material coordinate $X$ on some component $i$, its corresponding displacement can be written as $u(X) = A_i(X) {u}$, where $A_i(X)$ is the interpolation kernel for $X$ on component $i$, and ${u}$ is the concatenation of all nodal displacements from all components. So the above equality constraint for material coordinate $X$ on components $i$ and $j$ can be written as a linear constraint $A_i(X){u} - A_j(X){u}=0$. Every such equality constraint can be abstracted into a linear constraint $Hu = 0$. Likewise, since wing folding is driven by the transitioning of the central bistable mechanism, its boundary condition is given by the rotation of the arm from the bistable structure as a displacement control constraint. Therefore, the equilibrium equation we solve is as follows: 
\begin{equation}
\begin{split}
    \min_{u}\ e(u) \quad \text{s.t.} \quad  &u_i = u^*, \ \
                     H\widehat{u_i} = 0
\end{split}
\label{eq:joint_equilibrium}
\end{equation}
where we assume the displacement-controlled port is far from the joint so that matrix $H$ is full-rank. This optimization problem can also be solved using augmented Lagrangian. 

\begin{figure}
    \centering
    \includegraphics[width=0.45\textwidth]{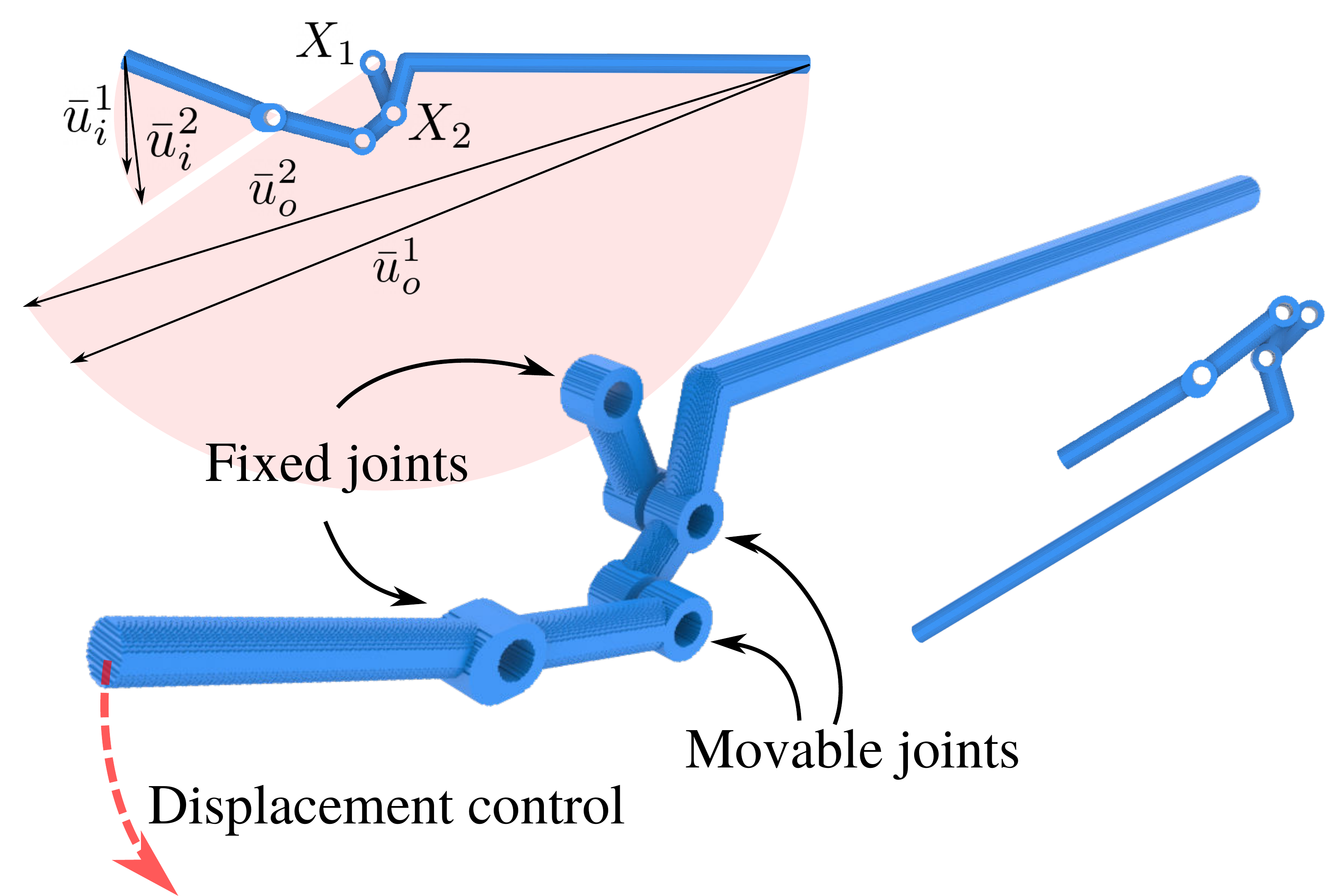}
    \caption{Optimized configuration. The deformation state of $u^2$ is illustrate on the bottom right corner. Top left shows how we choose the boundary condition and the objective.}
    \label{fig:initial_wing}
    \vspace{-0.5cm}
\end{figure}

We use three components and two joints in the design, where each joint connects two components (Fig.  \ref{fig:initial_wing}). This configuration choice is inspired by an earlier observation that three solid areas connected by two low-density joints will be formed when we only optimize over one single piece. One fixed joint is introduced to assist the rotational mechanism and another one serves as the rotation center of the arm. The formulation of the final wing optimization problem is then
\begin{equation}
\begin{split}
    &\min_{X_1, X_2} ||u^1_o - \bar{u}^{1}_o||_2^2 + ||u^2_o - \bar{u}^{2}_o||_2^2 + \alpha e(u^2)\\
    \text{s.t.}\ \ \ & ||(X_1 + u^2(X_1)) - (X_2 + u^2(X_2))|| \ge s_1\\
    & X_2^x + (u^2)^x(X_2) \ge s_2, \ \ X_1^y + (u^2)^y(X_1) \le s_3
\end{split}
\label{eq:wing_design}
\end{equation}
where $X_1, X_2$ are the two joints' material coordinates, $o$ indices the wing tip, and $\alpha = 10$ controls the weighting between objectives. During forward motion, the transition from the quadrotor mode to the fixed-wing mode is achieved by pulling the central plate followed by releasing it (Section \ref{sec:validation}), it will experience a larger deformation (with displacement field $u^2$) than the deformation at the second equilibrium (with displacement field $u^1$). Therefore we need to control this more deformed state as well. Specifically, $u^1$ and $u^2$ are both solved using Eq. \ref{eq:joint_equilibrium} under the same joint displacement equality constraints but different displacement controls, $u^1_i = \bar{u}^1_i$ and $u^2_i = \bar{u}^2_i$. We also minimize the compliance to ensure that no energy is stored in the structure. The first constraint is used to sufficiently separate the two movable joints for easier fabrication. The other two constraints prevent joints from colliding with the bounding boxes of other parts.

The terms in the objective and the constraints can all be abstracted into $L(u, \mathbf{X})$ that depends on the mesh displacement and joint positions, where $u$ also implicitly depends on $\mathbf{X}$ (the concatenation of the material coordinates of all movable joints to be optimized) through Eq. \ref{eq:joint_equilibrium}. The derivative of $L$ is given by
$
    \frac{dL}{d\textbf{X}} = \frac{\partial L}{\partial X} + \left[\frac{d\widehat{u_i}}{d\textbf{X}}\right]^T \frac{\partial L}{\partial \widehat{u_i}}
$,
where $\frac{d\widehat{u_i}}{d\textbf{X}}$ can be acquired by differentiating the equilibrium governing equation. The joint displacement constraints involve degrees of freedom that cannot be eliminated, thus we solve the governing equation using Lagrange multipliers 
and obtain $\frac{d\widehat{u_i}}{d\textbf{X}}$, and then compute $\frac{dL}{dX}$ for solving the optimization problem in Eq. \ref{eq:wing_design} with MMA.

In Fig. \ref{fig:initial_wing}, the final optimized wing is shown, the design domain and boundary condition is illustrated at the top left corner and its deformed state is illustrated at the bottom right corner. Note that the optimization is done in 2D with point joints. But the mechanism works in the same way when the point joints are extruded to be segments. The visualization is exactly how we assemble the wing parts in reality. We simplify the arm as a straight bar and control the displacement on the tip to simulate the rotation of the arm, which will be removed at fabrication and other parts will be attached to the bistable mechanism. The rotation angle of the arm at the second equilibrium and the maximal-deformation state is $0.83$ and $0.93$, where $\bar{u}^1_i$ and $\bar{u}^2_i$ are set accordingly. The auxiliary fixed joint is at (35 mm, 20 mm) w.r.t. the fixed joint of the arm. The rotation of the wing tip is assumed to be around (55 mm, 20 mm). $\bar{u}^1_o$ and $\bar{u}^2_o$ are determined when the rotation angles are $2.45$ and $2.46$ respectively. The rotation of the wing does not need to be precise, since we optimize with $L_2$ norm. $\bar{u}^1_o$ and $\bar{u}^2_o$ only provide a guide for the optimization, and we can tune the rotation angles of the wing tip a little to adjust the position of the wing blade.

\subsection{Validation through Forward Elastodynamic Simulations}
\label{sec:validation}

The computational procedures described above for automatic designs are based on quasi-static approximations. To more reliably predict whether the designed system can function properly in practice, the whole system needs to be tested with dynamic forward simulations.

We use implicit MPM \cite{wang2020hierarchical} to perform the elastodynamic simulations. At each time step, we execute backward Euler time integration (taking step size $\Delta t$) with lagged Rayleigh damping (system matrix only) from the last time step, where the 
nonlinear system can be reformulated into an incremental potential minimization problem \cite{gast2015optimization,li2019decomposed}. Considering the assembly constraints, we solve
\begin{equation*}
\begin{split}
    \min_{\Delta x} & \frac{1}{2}||\Delta x - \widetilde{\Delta x}||_M^2 + \frac{\gamma}{2\Delta t} ||\Delta x||_{K^n}^2 + \Delta t^2 (e(x^{n+1}) - \Delta x^T f_{ext}^n)\\
    & \text{s.t.} \quad  Hu = 0,
\end{split}
\end{equation*}
where $K^n = \frac{\partial^2 e}{dx^2}(x^n)$, $M$ is the mass matrix, $e$ is the elastic potential,  $f_{ext}$ is the external force, $\gamma$ is the damping coefficient, $v^{n,n+1}$ and $x^{n,n+1}$ denote velocities and positions from the known previous ($n$) and the unknown current ($n+1$) time steps, $\widetilde{\Delta x} = v^n\Delta t + g \Delta t^2$, and $||x||_A^2$ represents $x^T A x$. The above optimization is solved by augmented Lagrangian with projected Newton and non-invertible line search \cite{wang2020hierarchical,Li2020IPC}. For stability we adopt a total Lagrangian formulation for tracking the deformation \cite{de2020total}. 

\begin{figure}
    \centering
    \includegraphics[width=0.45\textwidth]{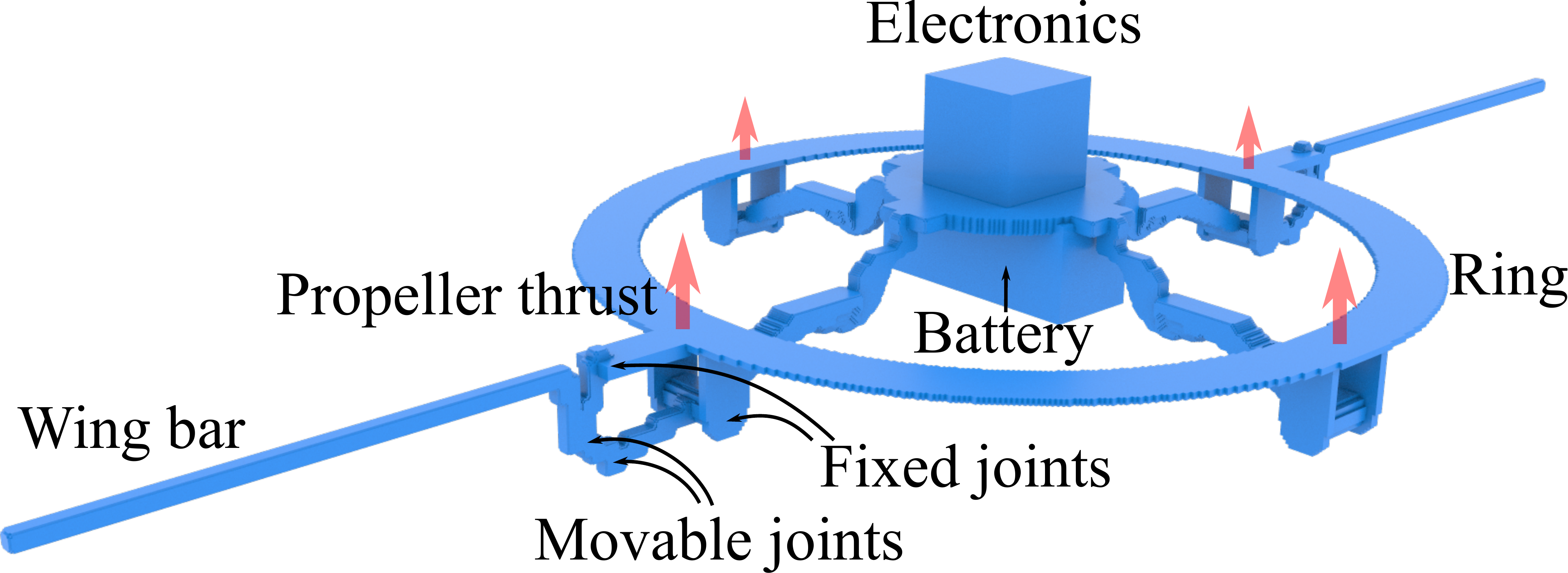}
    \caption{Final design (the simulated HAV system).}
    \label{fig:full_hav}
    \vspace{-0.5cm}
\end{figure}

\begin{figure}
    \centering
    \includegraphics[width=0.45\textwidth]{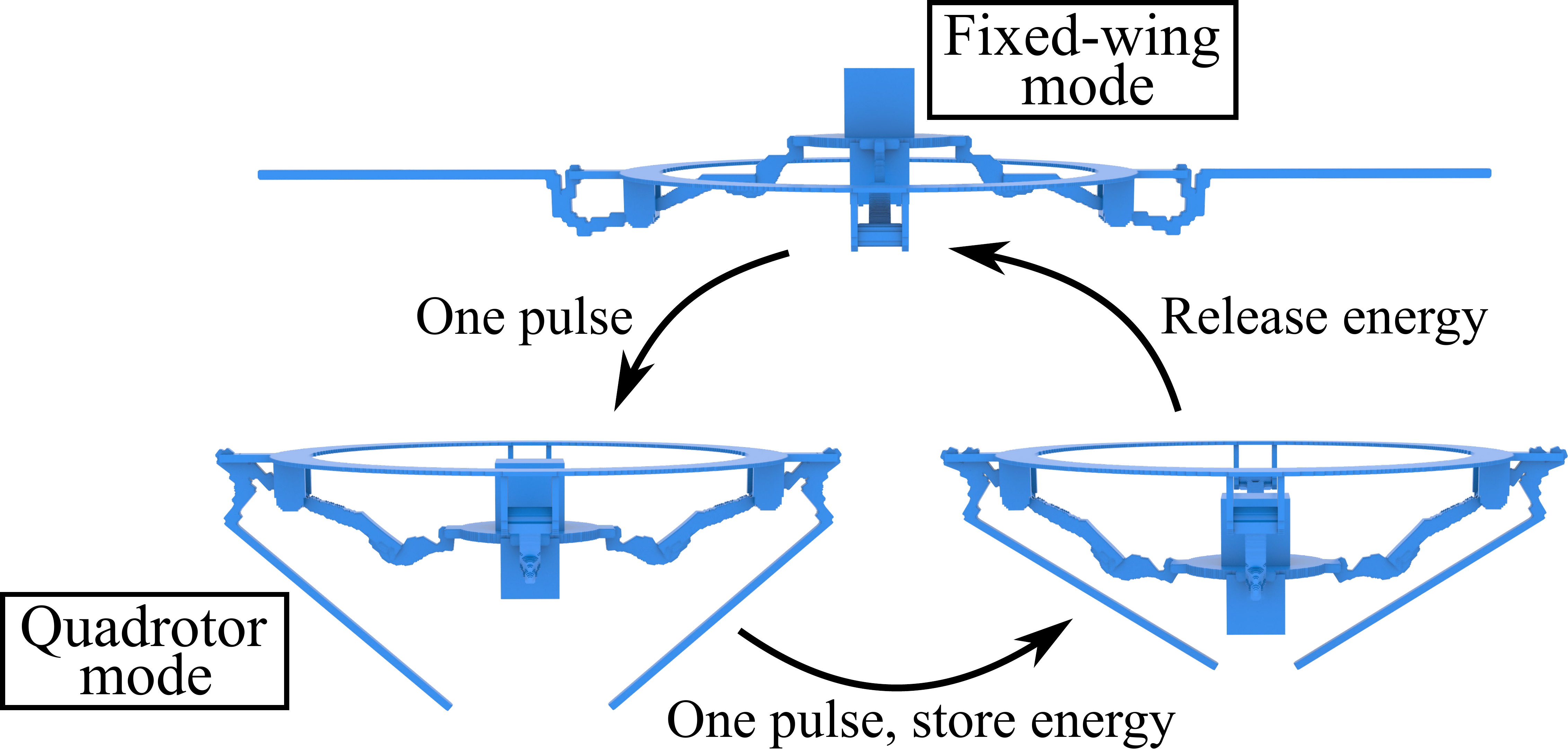}
    \caption{Theoretical state transition procedures.}
    \label{fig:transition}
    \vspace{-0.5cm}
\end{figure}

An assembled HAV is visualized in Fig. \ref{fig:full_hav}. The electronics and the battery are represented with boxes to simulate their inertia effects only. There are four propellers which are simulated as four external forces applied on the corresponding positions. Mode transition of the HAV relies on the inertia of the central mass. The theoretical transition procedure is illustrated in Fig. \ref{fig:transition}. It is easy to transit from the fixed-wing mode to the quadrotor mode: when the ring is accelerated suddenly, the fictitious inertial force will drag the central mass to the second equilibrium. However, it is not as straightforward to transit from the quadrotor mode to the fixed-wing mode since the propellers can only output unidirectional forces, which implies that the inertial force on the central mass is always pointing from the first equilibrium to the second equilibrium. Through experiments we discover a solution that utilizes the inertial force to temporarily store energy in the arm. After the propellers stop outputting forces, this energy is released to bounce the central mass from the second equilibrium towards the first equilibrium. The simulations are compared with real experiments in the next section to demonstrate the efficacy of our approach.

%% file: 4_evaluation.tex
\section{FABRICATION AND EVALUATION}
\subsection{Prototype}

We build the HAV prototype in Fig. \ref{fig:quad_proto} using the electronics and propellers from an ARRIS X220 V2 5" FPV Racing Drone \cite{Arrisx220}. The ARRIS drone has four 5" propellers placed 220m apart along the diagonal, but we increase the diameter to 250mm for our HAV to allow for motion of the electronics housing. The combination of ARRIS X2206 2450kv brushless motors and Dalprop T5045C high efficiency propellers produces 8.9N of thrust from each propeller when operating at 12V \cite{Motor}. The drone is powered by a 4S 1500mAh 100C LiPo battery. 
and can be remote controlled via radio signal. 
For experimental validation, we  use the provided remote control and communicate over radio using the Radiolink AT9S. 
The total mass of the off-the-shelf system is 490g including the battery. At 783g, our  prototype is not optimized for weight but the HAV has been shown in simulation and in testing to still have sufficient thrust.

For the switching mechanism, we 3D print 4 arms from TPU filament on a Makergear M3-ID 3D printer. Each arm replaces an arm of the ARRIS frame and connects from the housing to a ring laser cut out of 1/4$^{\prime\prime}$ thick acrylic sheet. To allow the arm to rotate and deform, it is attached to an cylindrical axel 20mm below the ring. The motors mounted to the top of the ring actuate the propellers.

We 3D print the 3 components of the wing from the topology optimization out of PLA filament. The long leading edge of the wing mechanism is constructed from a 8mm diameter carbon fiber rod. The second fixed joint is 3D printed as an extension to the connector holding the first fixed joint. We also add a rigid 3D printed bar extending into the ring that serves as an anchor for the wing surface. 

The wing surface is a folded arc segmented into 4 panels and fabricated out of a 0.005$^{\prime\prime}$ thick PET film (See Fig.~\ref{fig:quad_proto}. Two thin wing ribs cut from 1/8$^{\prime\prime}$ thick PETG rotate about $X_2$ and are attached to the second and third panels to guide the folding behavior \ref{fig:initial_wing}. We determined that for proper folding action, the central fold needs to be biased towards the folded state, which we accomplish by sewing this fold. 

The resulting prototype is bistable and able to support the weight of the battery against gravity, as predicted by Fig.~\ref{fig:two-stables} when in fixed wing mode, which will allow for a high climbing angle for the fixed wing.
When the bistable mechanism snaps through, the wing surface is able to collapse  and fold out as expected.

\subsection{1D testing without wing}
We mount the HAV to the end of a low friction boom with an arm length of 0.987m. 
The HAV was mounted at the end of the boom such that the thrust of the propeller is in the tangential direction. The setup was placed in a Vicon motion capture system to allow tracking of the boom's rotation. A GoPro Hero 8 mounted on the arm of the boom records the motion of the bistable mechanism.

\begin{figure}
    \centering
    \includegraphics[width=0.45\textwidth]{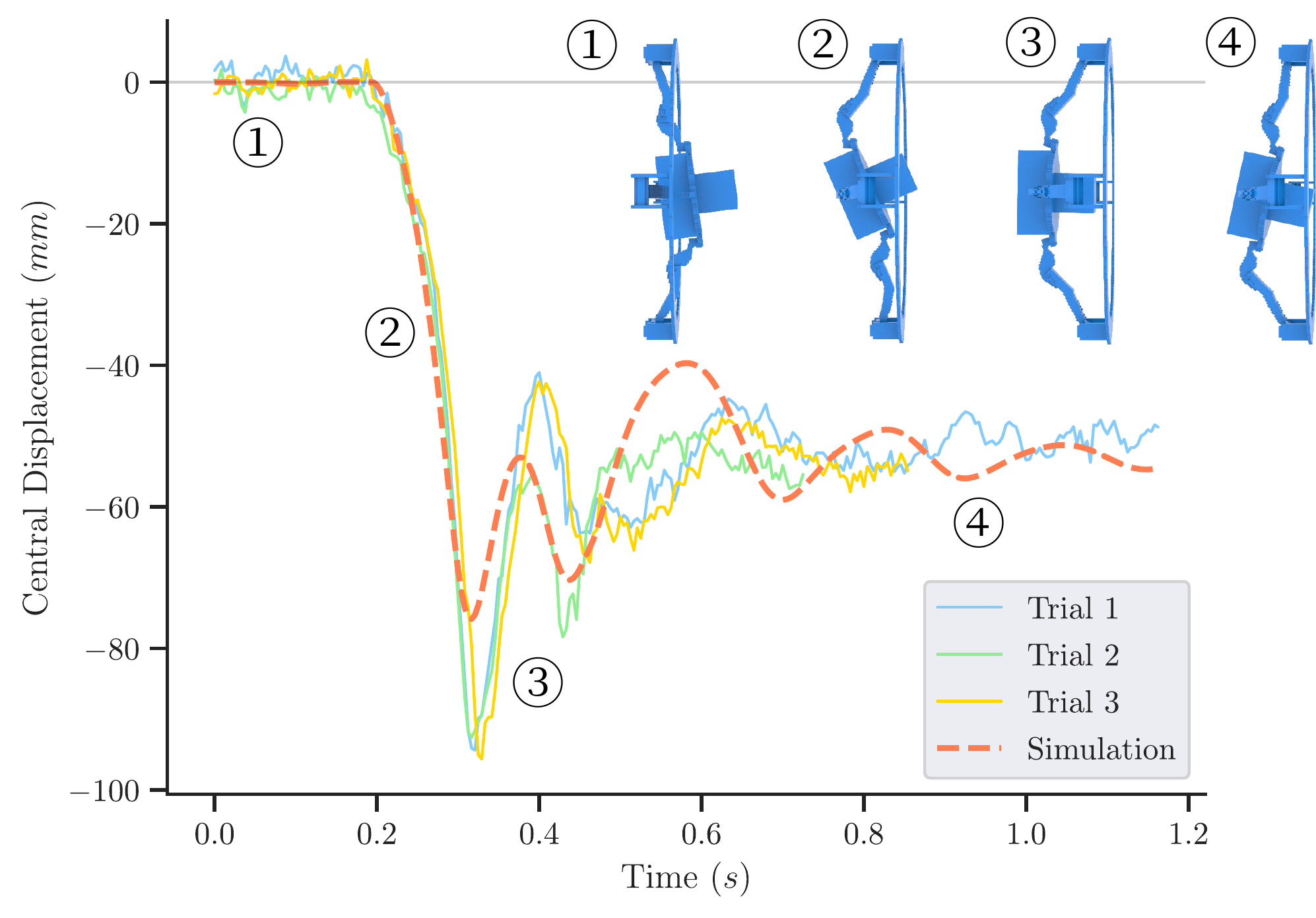}
    \caption{F to Q comparison}
    \label{fig:f2q_no_wing_comparison}
\end{figure}

\begin{figure}
    \centering
    \includegraphics[width=0.45\textwidth]{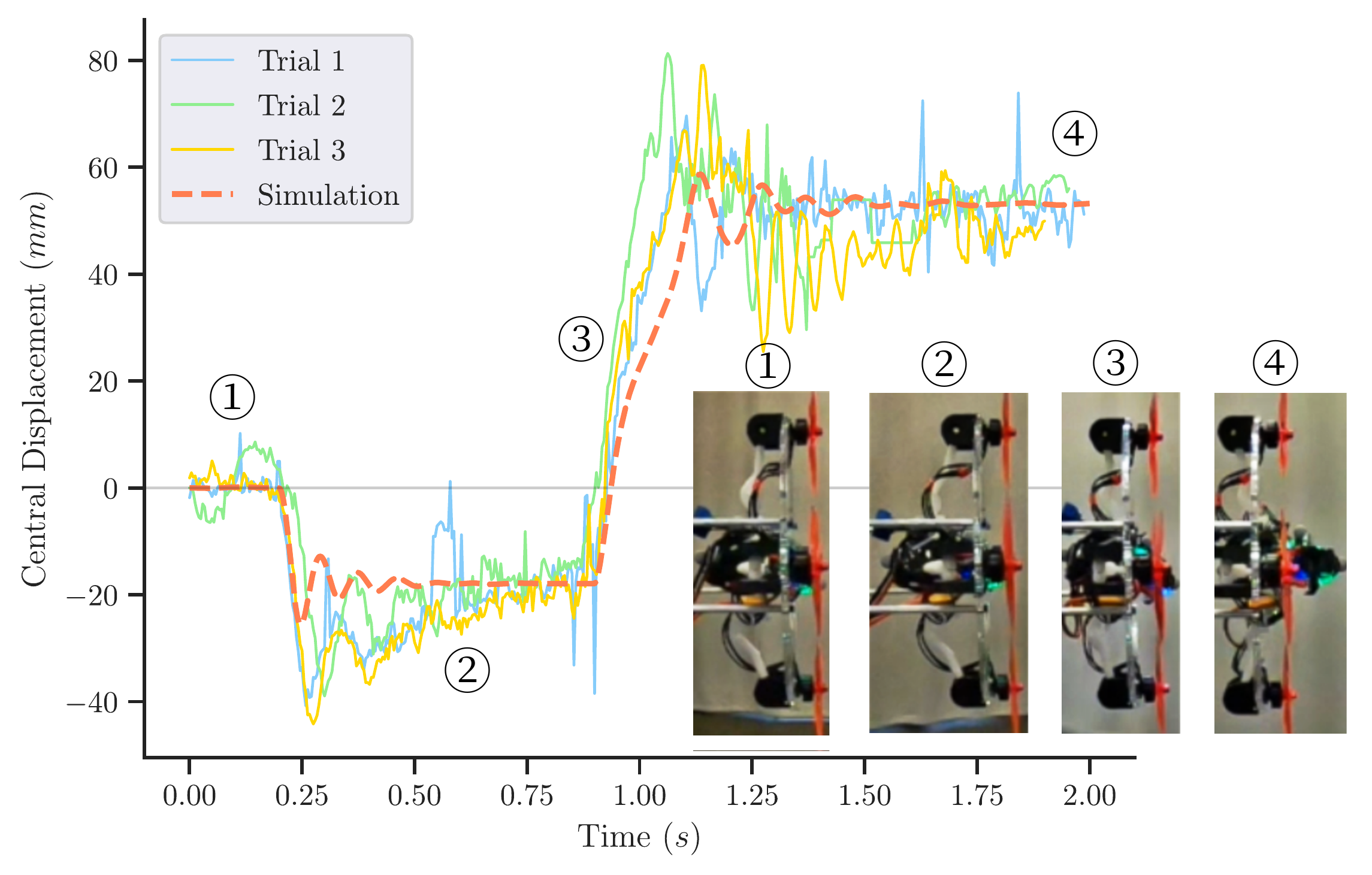}
    \caption{Q to F comparison}
    \label{fig:q2f_no_wing_comparison}
    \vspace{-0.5cm}
\end{figure}

We performed 3 trials of the quadrotor to fixed wing transition (Q to F) and 3 trials of the fixed wing to quadrotor (F to Q).
Each trial began with the HAV at rest.
The F to Q transition consists of one short, mid range pulse, while the Q to F transition consists of one long full thrust pulse, held for approximately 0.75s. Despite variability in control input due to manual operation of the HAV, the switching is reliable and the GoPro footage reveals that the central displacement of the bistable mechanism is similar to the predicted behavior of the simulation for both F to Q (Fig.~\ref{fig:f2q_no_wing_comparison}) and Q to F (Fig.~\ref{fig:q2f_no_wing_comparison}).

\subsection{1D testing with wing}
\begin{figure}
    \centering
    \includegraphics[width=0.35\textwidth]{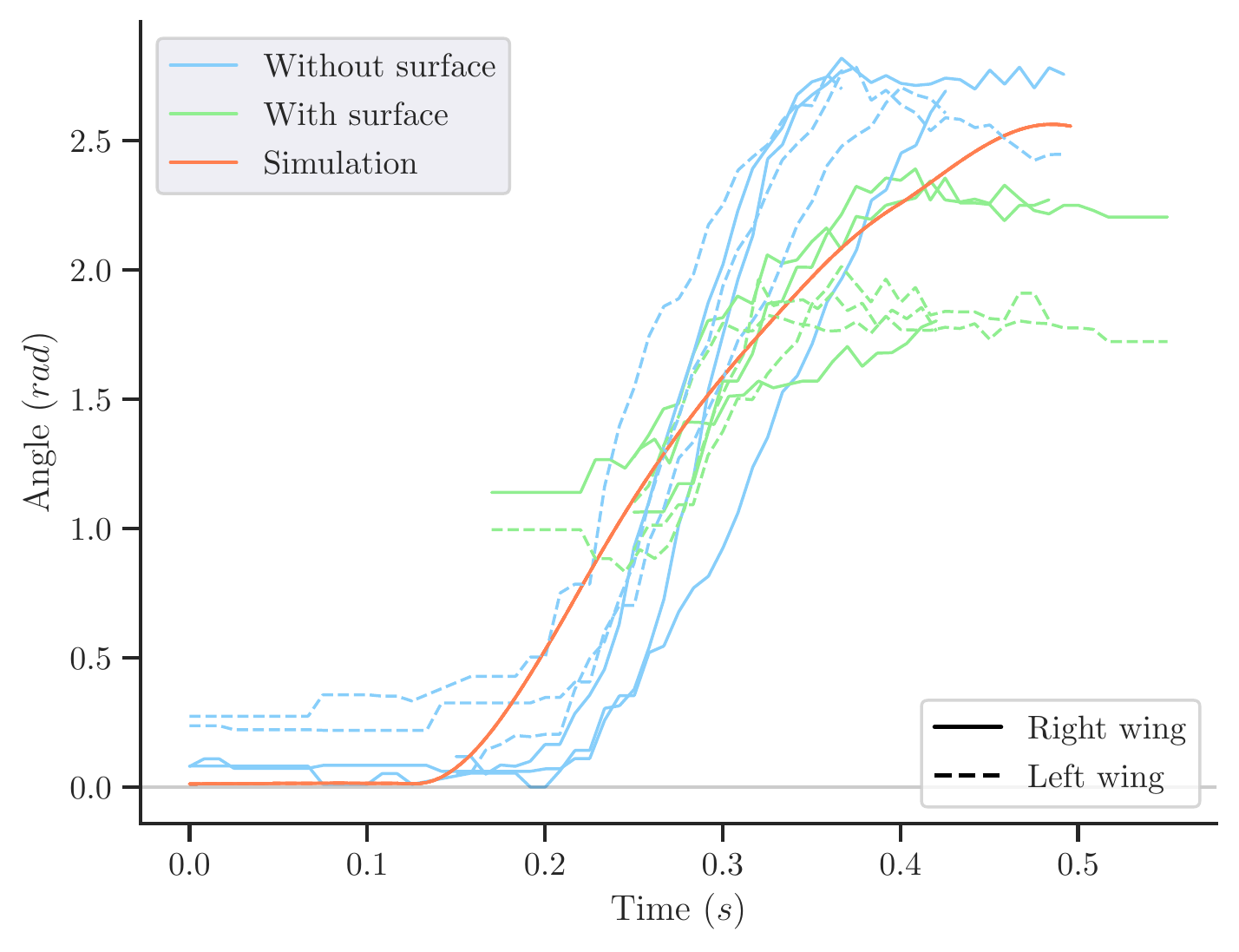}
    \caption{F to Q 1D wing angle comparison}
    \label{fig:f2q_wing_off_boom}
\end{figure}

\begin{figure}
    \centering
    \includegraphics[width=0.35\textwidth]{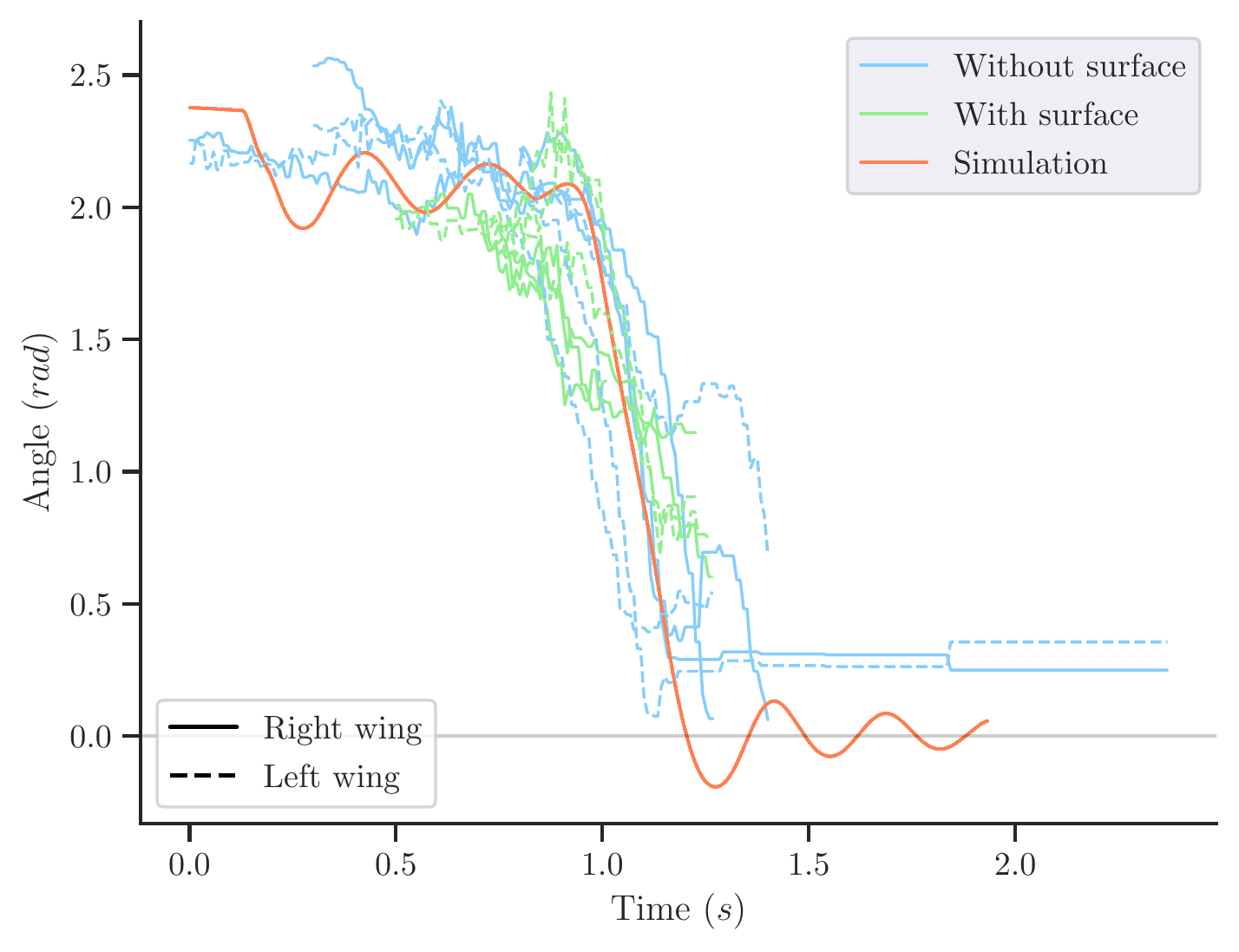}
    \caption{Q to F 1D wing angle comparison. An angle of zero corresponds to the fixed wing mode}
    \label{fig:q2f_wing_off_boom}
    \vspace{-0.5cm}
\end{figure}

When the HAV is constrained to rotate on a boom, the centripetal force dominates the rotation of the wing, causing both wing tips to rotate to the outside of the circle. Thus, performing boom tests with the wing did not provide useful information about the performance of the HAV. 
Instead, due to limited access to a large space and safety concerns with flying the untested prototype in a straight line, we conducted wing tests applying accelerations to the HAV prototype by hand. 
Grasping the ring of the HAV with the plane of the wing perpendicular to gravity, the experimenter manually exerted a pulsed force similar in profile to that measured during boom experiments. 
A GoPro fastened to the ceiling recorded the experiment at 120fps, and  the angle of the each wing was extracted. 
In all cases, the vehicle was able to successfully transition from Q to F and from F to Q, despite variability in the ``control inputs.''
Compared to the simulation, where wing deployment was symmetric, 
imbalances in the frictional forces caused small differences in wing fold-out angle (Fig. \ref{fig:q2f_wing_off_boom},\ref{fig:f2q_wing_off_boom}). 
When the wing surface was added, further deviation occurred since the folded wings added extra resistance to the foldout mechanism and prevented the mechanism from reaching its extreme angles.



%% file: 5_discussion.tex
\section{DISCUSSION}
We have demonstrated a novel design for a morphing HAV that leverages a bistable mechanism and vehicle accelerations to change flight modes.
A topology optimization approach successfully generates a bistable mechanism with appropriate snap-through force that is low enough for the HAV's motors to trigger snap-through but high enough to prevent accidental mode switching. We showed that reliable mode switching 
is possible on the physically constructed device, even under variable acceleration inputs. We also demonstrate topology optimization of a folding wing mechanism driven by the motion of the bistable mechanism. 


Additional design iterations will likely need to be performed to create an efficient HAV design.
During experiments, we discovered that since the connection of the TPU arm to the fixed joint is thin, sometimes the fabricated prototype deforms here rather than rotating, and thus does not transmit the motion to the wing. 
We will iterate over the design of this connection so that there is no compliance at the axel.
Further, the wing surface has not yet been optimized.

Future work includes progressing to more 3D flight testing and the development of a controller for the switching movement. To go from quadrotor to fixed wing will involve a pitch forward motion, which has been qualitatively observed to aid in the Q to F transition. Similarly, the fixed wing will need to pitch up when it transforms to quadrotor mode, which will allow for a much smoother thrust application to cause the transition when it is combined with pitch.